\documentclass{article}
\usepackage{spconf,amsmath,graphicx}

\title{''You should probably read this'': Hedge Detection in Text}
\twoauthors
    {Denys Katerenchuk$^1$}
	{dkaterenchuk@gradcenter.cuny.edu\\
	$^1$The Graduate Center, CUNY\\
	Department of Computer Science\\
	New York, USA}
	{Rivka Levitan$^{1,2}$}
	{levitan@sci.brooklyn.cuny.edu\\
	$^2$Brooklyn College, CUNY\\
	Department of Computer and Information Science\\
    Brooklyn, USA}
\begin{document}
%
\maketitle
\begin{abstract}
Humans express ideas, beliefs, and statements through language. The manner of expression can carry information indicating the author's degree of confidence in their statement. Understanding the certainty level of a claim is crucial in areas such as medicine, finance, engineering, and many others where errors can lead to disastrous results. In this work, we apply a joint model that leverages words and part-of-speech tags to improve hedge detection in text and achieve a new top score 
on the CoNLL-2010 Wikipedia corpus. 

\end{abstract}

\section{Introduction}
Imagine a situation where a doctor says to a patient \textit{''I think you need surgery immediately!''} Many will take this statement seriously without any additional considerations. However, the phrase \textit{''I think''} signals uncertainty in the diagnosis. Identifying these signs of uncertain claims, known as hedges, can prevent cases of malpractice and save lives. For this reason, uncertainty detection is an important problem in medicine, finance, engineering, and other high-risk fields. In this work, we explore CoNLL-2010 shared task data \cite{farkas2010conll} to improve current methods of hedge detection.



The degree of certainty in language was first introduced by G. Lakoff \cite{lakoff1973lexicography}. His work introduces the concept of hedges, which are linguistic devices that are used in conversations to indicate the degree of belief. Hedge phrases can be expressed through modal verbs (''could'', ''might'', etc.), peacock expressions (''very likely'', ''everyone'', ''I think'', etc.), and weasel words (''some believe'', ''clearly'', etc.). These expressions are context-dependent and the presence of these in a sentence doesn't indicate uncertainty. Hence, this problem is difficult and advanced methods are required to identify uncertain claims.

In this work, we focus on detection sentences that contain hedges. We explore different neural network architectures and present a joint model problem formulation to include part-of-speech tags to improve current results on the CoNLL-2010 Wikipedia dataset. While the data used is based on Wikipedia, this work can be applied to speech as a downstream task and other similar problems in the NLP domain. 
The main contributions of this work are 1) an analysis of various neural network architectures and their performance, 2) a model formulation for including part-of-speech information in the input, and 3) a new top score on the CoNLL-2010 Wikipedia dataset.

\section{Related Work}
Hedge detection has attracted much attention in recent years. The early work relies on shallow linguistic features such as n-grams, manually crafted hedge-word lists, and word-scoring functions \cite{ganter2009finding}. M. Georgescul \cite{georgescul2010hedgehop} showed the highest result during the CoNLL-2010 competition, achieving the F1 score of 60.2 using an SVM model \cite{vapnik1998support} with bag-of-words (BoW) features. The general early approach to this problem is a combination of word and context analysis with BoW and part-of-speech (POS) tags represented as various size n-grams \cite{choi2012hedge}. 

With the popularity of neural networks (NN), Patel et al. \cite{patelmodeling} applied a convolutional neural network (CNN) \cite{kim2014convolutional} to the problem of ambiguity detection in legal literature and showed that this model outperforms SVM models. The addition of attention \cite{chorowski2015attention} to the CNN model achieved an F1 score of 67.52 on the CoNLL-2010 Wikipedia data \cite{adel2017exploring}. This work produced the current best result on the Wikipedia data\footnote{Han et al. \cite{han2020attention} published new results while this paper was under review. }. The work proposes attention formulations that take advantage of attention weights learned from the sentences and provided cue phrases. Their work and the F1 score of 67.52 is used as the baseline in our paper.

In recent years, the popularity of transformer architecture \cite{vaswani2017attention} and transformer-based language models \cite{devlin2019bert,radford2019language} led to improvements in NLP tasks, including uncertainty detection. Sinha et al. \cite{sinha2020relation} showed that the fine-tuned BERT model outperformed Tree-LSTM and CNN based model on two out of three corpora, including the BioScope corpus from the CoNLL-2010 challenge. However, the authors did not report the score on the Wikipedia dataset.

Motivated by the early works and POS tags' use, this paper makes an effort to leverage a joint word and POS model formulation to achieve a new high score on the CoNLL-2010 Wikipedia dataset.
The performance improvements gained by applying transformer-based language models to the BioScope corpus further motivate our exploration of more complex models on the Wikipedia corpus. 


\section{Data}
The CoNLL-2010 Wikipedia dataset was released by \cite{farkas2010conll} as a part of the CoNLL-2010 challenge. The task provides data that is derived from two datasets: biological scientific articles
and articles from Wikipedia. Both datasets were manually annotated by two independent linguists who were directed to identify \textit{cues}, which are expressed through the use of auxiliaries, hedge verbs with speculations, adjectives, adverbs, conjunctions, and weasel words (for the Wikipedia data). A third linguist was invited to evaluate the differences. Each sentence is labeled as certain or uncertain, and phrases that identify uncertainty are annotated as cues. The majority (77.26\%) of the sentences are certain; as a result, the data is unbalanced. Hence, the evaluation results are reported as the F1-score.

While the CoNLL-2010 shared task provides data from both BioScope and Wikipedia, this work focuses solely on the Wikipedia data for the following reasons: 1) the language on Wikipedia is more general and not limited to any domains. Hence, the methods applied here are expected to be generalized and applicable to other similar problems. 2) the Wikipedia data contains annotations for weasel words, which are statements of unsupported claims. These weasel words make the uncertainty detection problem more difficult, which can be noted from the current best scores: F1 67.52 for Wikipedia and F1 86.22 for the BioScope data. For these reasons, all the experiments in this paper are focused on Wikipedia data.

The CoNLL-2010 Wikipedia dataset contains 11,110 training and 9,634 evaluation sentences. 10\% of the training data is randomly selected as the development data. The models are tested on the evaluation data only used after being optimized on the development dataset. The corpus is pre-processed by tokenizing the words and stripping any punctuation that is web noise or doesn't contribute to the sentence semantics (such as \textit{''<>/-*''}). The dataset is POS annotated with spaCy \cite{spacy2}, resulting in two parallel strings containing words and POS tags, respectively.  Given that the mean sentence length is 22.5 words, the sentence length is limited to 64 words.

\section{Methods}
 In this work, we conduct experiments to improve current results on the CoNLL-2010 Wikipedia corpus. The training data contains only 11,110 sentences, and 10\% (1,111) of that data is used as a development set for model selection. The training dataset with 9,999 data points is very small for today's standards, presenting several challenges.
 We survey various word embedding models and neural network architectures to find the best model for this task.
 Furthermore, we look at POS tags for additional information on sentence structure. This POS tag information is used in a joint model approach to improve performance on CoNLL-2010 Wikipedia data. Finally, we analyze the models and present the final results on the evaluation dataset.

\subsection{Pre-trained Word Embedding models}

Pre-trained word embeddings (WE) \cite{mikolov2013distributed} are hard to beat due to the sheer size of the data on which they are trained. However, the mismatch between their training data and the application domain can affect performance \cite{gu2020domain}. We train a custom WE model on 1 GB of randomly selected Wikipedia data, using the FastText Algorithm \cite{bojanowski2017enriching}. 
This trained model (Wiki 1G) has 256 dimensions and much smaller in size compared to available pre-trained models.
We compare our in-domain model with several available pre-trained models, using a single layer bi-directional LSTM model \cite{hochreiter1997long} to choose the ones that perform best on this task and data. Table \ref{tab:word_emb} shows the results of each word embedding language model on the development set. The score represents the mean of 10 samples every 10 iterations starting at 400 iterations. The mean score should be a more robust representation of performance.

\begin{table}[htp]
\centering
\begin{tabular}{|l|l|}
\hline
Word Embed. Models & F1 score \\
\hline
GoogleNews 300d & 60.34 \\ 
GloVe 100d & 55.74 \\  
GloVe 300d 6B & 63.12 \\  
GloVe 300d 840B & 62.41 \\  
FastText 1M &  62.09 \\ 
FastText 2M &  63.57 \\ 
Wiki 1Gb & 61.99  \\ 
\hline
\end{tabular}
\caption{Word Embedding Models}
\label{tab:word_emb}
\end{table}

The results in Table \ref{tab:word_emb} show that FastText 2M and GloVe 6B language models work the best, outperforming the rest. Hence, future experiments are based on these two language models. However, we also include the custom-trained Wiki 1G. This model is small in size and performs comparably to the larger models, making it a good choice for prototyping.

\subsection{Neural Networks}
\label{sec:neural_networks}

To find the best-performing model for this task, we evaluate the CNN \cite{kim2014convolutional}, GRU \cite{chung2014empirical}, and LSTM \cite{hochreiter1997long} architectures. These architectures with an attention layer have shown the top F1 score of 67.52 on the Wikipedia data \cite{adel2017exploring}. Additionally, we explore multi-head attention model architecture with the positional encoding \cite{vaswani2017attention}. The attention architectures are used as a standalone two-layer transformer model and as an additional layer on top of the base architecture. The initial hyperparameters are chosen from earlier research in this domain and tuned during the training. Thus, each model is trained with the following hyperparameters: 64 - RNN hidden units, 64 - batch size, and 0.5 dropout rate at each layer. Each model has a two-layer architecture with an added transformer layer where specified. The final layer is a dense layer with a single neuron as an output. We find a two-layer architecture optimal in our experiments. The models are trained with a stochastic gradient descent optimizer on the training dataset and evaluated on the development dataset. We apply the class weights during the training to mitigate the bias towards the majority class. This is done by calculating the class distributions in the training dataset and multiplying it by the class error rate during the training.
In order to get the best results and avoid finding a ''lucky'' configuration, the models are trained until they reach optimal performance, which is indicated by the development error evening out without diverging from the training error. At this point, the results are sampled for 10 more iterations and the mean score on the development dataset is reported in Table \ref{tab:nn_models}.


\begin{table*}[!h]
\centering
\begin{tabular}{|l|lll|ll|}
\hline
 NN Models & Wiki 1G & GloVe & FastText & Mean & STD \\ 
\hline
CNN & 57.38 & 58.34 & 59.36 & 58.36 & 0.99 \\  
GRU & 62.07 & 64.23 & 65.04 & \textbf{63.78} & 1.54 \\  
LSTM & 64.35 & 62.14 & 63.91 & \textbf{63.47} & 1.17 \\  
Transformer  & 57.18 & 57.91 & 54.74 & 56.61 & 1.66  \\ 
CNN + At & 56.86 & 58.22 & 62.79 & 59.29 & 3.11 \\  
GRU + At & 62.13 & 63.33 & 65.4 & \textbf{63.62} & 1.65 \\  
LSTM + At & 59.37 & 60.11 & 64.26 & \textbf{61.25} & 2.64 \\  
\hline
\end{tabular}
\caption{Neural Network Models}
\label{tab:nn_models}
\end{table*}

The results in Table \ref{tab:nn_models} show that 4 RNN-based architectures show the best performance, with the GRU-based models performing better on average. We believe that is because GRUs are simpler models. Considering that the training set contains only 9999 sentences, complex models overfit on this task. This could be the reason we did not see greater performance from CNN or transformer models. The pretrained Roberta \cite{liu2019roberta} model tuned on hedge detection produces 60.26 F1 before overfitting on the task.
Hence, LSTM and GRU based networks are chosen for the rest of the experiments.



\begin{figure}[htp]
    \centering
    \includegraphics[width=5cm]{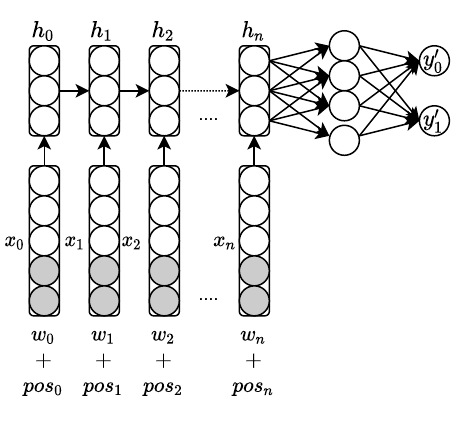}
    \caption{Word \& POS Input Joint Model}
    \label{pitc:word_pos_input}
\end{figure}
\begin{figure}[htp]
    \centering
    \includegraphics[width=5cm]{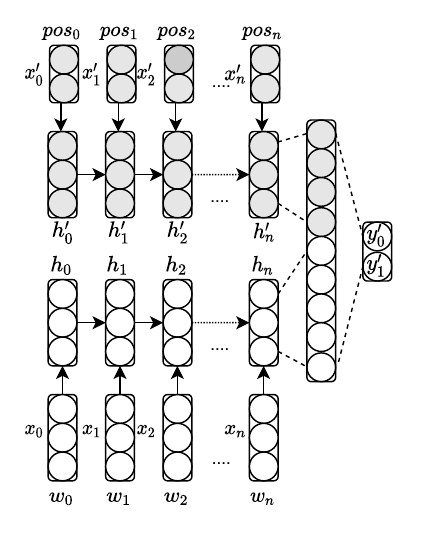}
    \caption{Latent Space Joint Model}
    \label{pict:joint_model}
\end{figure}

\subsection{POS Models}
 Early research on hedge detection relied on both words and POS tags \cite{choi2012hedge}. However, POS tags have not been included in recent work on this task. Motivated by the early works in this domain, we propose to leverage additional POS tag information to improve uncertainty detection.
 We train 4 models: LSTM, GRU, and their variants with an attention layer on POS tags to classify uncertain sentences. 
 The POS tag embeddings are learned during the training as a part of the model. We chose to encode the tags in 8-dimensional vectors.
 All four models perform similarly with the results falling in the range between 47.64 - 48.9 F1 score. The LSTM architecture with an attention layer achieves the top score. Even though the results are in a close range, we believe that the more complex LSTM architecture can better capture POS tag sentence information. 
 This result shows that POS tags contain a predictive signal that can be leveraged to improve performance.

\subsection{Word \& POS Joint Model}

Having demonstrated that POS tags contain a predictive signal, we evaluate two different ways to leverage POS information and combine them with WE-based hedge-prediction models. One way to include the POS information is to concatenate POS tag representations to word embeddings. This approach enhances the word representation by encoding not only the meaning but also its function in a sentence (Fig. \ref{pitc:word_pos_input}). Another approach is to have two separate networks for POS and word embeddings. These networks are independent and join together only in the last hidden layer. This last hidden layer is used as an input to the output dense layer for the final class prediction (Fig. \ref{pict:joint_model}). We hypothesize that the combination of word-GRU and POS-LSTM networks with the transformer layer should be the most predictive architecture that leverages two independent sentence representations.
The results are shown in Table \ref{tab:joint} with the top 3 scores highlighted.


\begin{table}[htp]
\centering
\begin{tabular}{|lll|ll|}
\hline
  & Wiki 1G & FastText & Mean & STD \\ 
\hline
\multicolumn{5}{|l|}{Joint Input} \\
\hline
GRU & 65.81 & 66.08 & \textbf{65.95} & 0.19 \\  
LSTM  & 61.69 & 66.14 & 63.92 & 3.15 \\  
GRU + At & 65.46 & 65.45 & \textbf{65.46} & 0.01 \\  
LSTM + At & 66.57 & 64.26 & 65.42 & 1.63 \\  
\hline
\multicolumn{5}{|l|}{Joint RNN + Attention Models}  \\
\hline
GRU & 64.42 & 64.54 & 64.48 & 0.08 \\  
LSTM& 64.22 & 62.47 & 63.35 & 1.24 \\  
GRU\&LSTM& 64.82  & 66.09 & \textbf{65.46} & 0.9 \\ 
\hline
\end{tabular}
\caption{Joint POS \& Word Models}
\label{tab:joint}
\end{table}

From Table \ref{tab:joint}, we can see that the addition of POS information improves the performance for Wiki 1Gb and FastText-based architectures compared to the results reported in Section \ref{sec:neural_networks}. We can see the improvements across different network architectures with GRU, GRU with attention, and the joint model of GRU and LSTM, both with attention, show top performance. The GloVe-based language models, however, did not show any improvements and were excluded from further experiments. 
At this point, we test the three best-performing models on the evaluation dataset and report the results.


\section{Results}

We narrow down the choice of models to three architectures to be used on the evaluation dataset. The architectures are joint input GRU and GRU+att, and joint latent space model of word GRU+att and POS LSTM+att. The main question is whether the addition of POS tags can help get better results on the evaluation dataset over the previous top F1 score of 67.52. 

First, we start by testing our hypothesis by training an LSTM model on POS tags. The result of this model on the evaluation is equal to a 55.67 F1 score. We find this score high considering that the model makes predictions only on POS tags and promising for our joint model formulation. Next, the joint models are evaluated on the test dataset.

\begin{table}[htp]
\centering
\begin{tabular}{|l|ll|ll|}
\hline
 & Wiki & FastText & Mean & STD \\ 
\hline

\multicolumn{5}{|l|}{Joint Input Models} \\
\hline
GRU & 68.25 & 67.69 & 67.97 & 0.4 \\  
GRU + At & 68.97 & 66.32 & 67.65 & 1.87 \\  
\hline
\multicolumn{5}{|l|}{Joint RNN + Attention Models} \\
\hline
GRU\&LSTM & \textbf{69.21} & \textbf{69.74} & \textbf{69.48} & 0.37 \\ 
\hline
\end{tabular}
\caption{Final Results on the Eval Dataset}
\label{tab:results}
\end{table}

From table \ref{tab:results} we can see that the joint model of word GRU+att and POS LSTM+att produces top scores across the two language models. Both scores outperform the current top result on this dataset and show that leveraging POS tag information in a joint model formulation can improve the results. The previous top score on this dataset \cite{adel2017exploring} introduces a novel concept of external attention where a scoring function measures each word's similarity with respect to uncertainty-indicating phrases. These phrases would need to be derived by a human from the training dataset. Our paper shows that POS tags' addition can achieve higher performance without a need for additional human annotations.

Furthermore, we test the performance of all models mentioned in the earlier sections on the evaluation dataset. To our surprise, a simple GRU model with custom-trained word embeddings (Wiki 1g) achieves the top score of 70.24. We hypothesize that this could be because the domain-specific language model can deliver better word representations and outperform general language models. Besides, a simpler GRU-based model architecture plays a role of regularization further improving the model performance on this small dataset. However, the results are dependent on custom word embeddings, which introduces complicity and variance.



\section{Conclusions and Future work}
In this work, we show that POS tags introduce additional information that can be used to improve results on hedge detection. We formulate the problem as a joint model of POS tags and words trained together. This joint model formulation achieves a high 69.74 F1 score and 70.24 F1 score with domain-specific word embeddings on the CoNLL-2010 Wikipedia dataset. This work introduces a joint POS and word approach that works well on hedge detection. In future work, we would like to continue experiments in this domain to answer questions about whether this joint model formulation works on different datasets as well as other problems in this domain. In addition, we would like to test if domain-specific word embedding performs well on other datasets. Lastly, transformers improve the performance of many NLP tasks, and further research in hedge detection applications is needed.


\newpage

\bibliography{hedge}

\begin{thebibliography}{10}

\bibitem{farkas2010conll}
Rich{\'a}rd Farkas, Veronika Vincze, Gy{\"o}rgy M{\'o}ra, J{\'a}nos Csirik, and Gy{\"o}rgy Szarvas,
\newblock ``The conll-2010 shared task: learning to detect hedges and their scope in natural language text,''
\newblock Association for Computational Linguistics, 2010.

\bibitem{lakoff1973lexicography}
George Lakoff,
\newblock ``Lexicography and generative grammar i: Hedges and meaning criteria,''
\newblock {\em Annals of the New York Academy of Sciences}, vol. 211, no. 1, pp. 144--153, 1973.

\bibitem{ganter2009finding}
Viola Ganter and Michael Strube,
\newblock ``Finding hedges by chasing weasels: Hedge detection using wikipedia tags and shallow linguistic features,''
\newblock in {\em Proceedings of the ACL-IJCNLP 2009 Conference Short Papers}, 2009, pp. 173--176.

\bibitem{georgescul2010hedgehop}
Maria Georgescul,
\newblock ``A hedgehop over a max-margin framework using hedge cues,''
\newblock in {\em Proceedings of the 14th International Conference on Computational Natural Language Learning: Shared Task}, 2010, pp. 26--31.

\bibitem{vapnik1998support}
Vladimir Vapnik,
\newblock ``The support vector method of function estimation,''
\newblock in {\em Nonlinear Modeling}, pp. 55--85. Springer, 1998.

\bibitem{choi2012hedge}
Eunsol Choi, Chenhao Tan, Lillian Lee, Cristian Danescu-Niculescu-Mizil, and Jennifer Spindel,
\newblock ``Hedge detection as a lens on framing in the gmo debates: a position paper,''
\newblock pp. 70--79, 2012.

\bibitem{patelmodeling}
Roma Patel and A.~Nenkova,
\newblock ``Modeling ambiguity in text: A corpus of legal literature,''
\newblock 2019.

\bibitem{kim2014convolutional}
Yoon Kim,
\newblock ``Convolutional neural networks for sentence classification,''
\newblock pp. 1746--1751, Oct. 2014.

\bibitem{chorowski2015attention}
Jan~K Chorowski, Dzmitry Bahdanau, Dmitriy Serdyuk, Kyunghyun Cho, and Yoshua Bengio,
\newblock ``Attention-based models for speech recognition,''
\newblock in {\em Advances in neural information processing systems}, 2015, pp. 577--585.

\bibitem{adel2017exploring}
Heike Adel and Hinrich Sch{\"u}tze,
\newblock ``Exploring different dimensions of attention for uncertainty detection,''
\newblock in {\em Proceedings of the 15th Conference of the European Chapter of the Association for Computational Linguistics: Volume 1, Long Papers}, 2017, pp. 22--34.

\bibitem{han2020attention}
Xu~Han, Binyang Li, and Zhuoran Wang,
\newblock ``An attention-based neural framework for uncertainty identification on social media texts,''
\newblock {\em Tsinghua Science and Technology}, vol. 1, 2020.

\bibitem{vaswani2017attention}
Ashish Vaswani, Noam Shazeer, Niki Parmar, Jakob Uszkoreit, Llion Jones, Aidan~N Gomez, {\L}ukasz Kaiser, and Illia Polosukhin,
\newblock ``Attention is all you need,''
\newblock in {\em Advances in neural information processing systems}, 2017, pp. 5998--6008.

\bibitem{devlin2019bert}
Jacob Devlin, Ming-Wei Chang, Kenton Lee, and Kristina Toutanova,
\newblock ``Bert: Pre-training of deep bidirectional transformers for language understanding,''
\newblock 2019.

\bibitem{radford2019language}
Alec Radford, Jeffrey Wu, Rewon Child, David Luan, Dario Amodei, and Ilya Sutskever,
\newblock ``Language models are unsupervised multitask learners,''
\newblock {\em OpenAI Blog}, vol. 1, no. 8, pp. 9, 2019.

\bibitem{sinha2020relation}
Manjira Sinha, Nilesh Agarwal, and Tirthankar Dasgupta,
\newblock ``Relation aware attention model for uncertainty detection in text,''
\newblock in {\em Proceedings of the ACM/IEEE Joint Conference on Digital Libraries in 2020}, 2020, pp. 437--440.

\bibitem{spacy2}
Matthew Honnibal, Ines Montani, Sofie Van~Landeghem, and Adriane Boyd,
\newblock ``{spaCy: Industrial-strength Natural Language Processing in Python},'' 2020.

\bibitem{mikolov2013distributed}
Tomas Mikolov, Ilya Sutskever, Kai Chen, Greg~S Corrado, and Jeff Dean,
\newblock ``Distributed representations of words and phrases and their compositionality,''
\newblock {\em Advances in neural information processing systems}, vol. 26, 2013.

\bibitem{gu2020domain}
Yu~Gu, Robert Tinn, Hao Cheng, Michael Lucas, Naoto Usuyama, Xiaodong Liu, Tristan Naumann, Jianfeng Gao, and Hoifung Poon,
\newblock ``Domain-specific language model pretraining for biomedical natural language processing,''
\newblock {\em arXiv preprint arXiv:2007.15779}, 2020.

\bibitem{bojanowski2017enriching}
Piotr Bojanowski, Edouard Grave, Armand Joulin, and Tomas Mikolov,
\newblock ``Enriching word vectors with subword information,''
\newblock {\em Transactions of the Association for Computational Linguistics}, vol. 5, pp. 135--146, 2017.

\bibitem{hochreiter1997long}
Sepp Hochreiter and J{\"u}rgen Schmidhuber,
\newblock ``Long short-term memory,''
\newblock {\em Neural computation}, vol. 9, no. 8, pp. 1735--1780, 1997.

\bibitem{chung2014empirical}
Junyoung Chung, Caglar Gulcehre, Kyunghyun Cho, and Yoshua Bengio,
\newblock ``Empirical evaluation of gated recurrent neural networks on sequence modeling,''
\newblock in {\em NIPS 2014 Workshop on Deep Learning, December 2014}, 2014.

\bibitem{liu2019roberta}
Yinhan Liu, Myle Ott, Naman Goyal, Jingfei Du, Mandar Joshi, Danqi Chen, Omer Levy, Mike Lewis, Luke Zettlemoyer, and Veselin Stoyanov,
\newblock ``Roberta: A robustly optimized bert pretraining approach,'' 2019,
\newblock cite arxiv:1907.11692.

\end{thebibliography}
\bibliographystyle{icassp_natbib}
\end{document}